\PassOptionsToPackage{table,xcdraw}{xcolor}

\documentclass[sigconf, nonacm, natbib=false]{acmart}
\PassOptionsToPackage{numbers, compress}{natbib}

\copyrightyear{2023}
\acmYear{2023}
\setcopyright{acmlicensed}\acmConference[CIKM '23]{Proceedings of the 32nd ACM International Conference on Information and Knowledge Management}{October 21--25, 2023}{Birmingham, United Kingdom}
\acmBooktitle{Proceedings of the 32nd ACM International Conference on Information and Knowledge Management (CIKM '23), October 21--25, 2023, Birmingham, United Kingdom}
\acmPrice{15.00}
\acmDOI{10.1145/3583780.3614875}
\acmISBN{979-8-4007-0124-5/23/10}

\RequirePackage[
  datamodel=acmdatamodel,
  style=acmnumeric,
]{biblatex}

\usepackage[utf8]{inputenc} 
\usepackage[T1]{fontenc}    
\usepackage{hyperref}       
\usepackage{url}            
\usepackage{booktabs}       
\usepackage{amsfonts}       
\usepackage{nicefrac}       
\usepackage{microtype}      
\usepackage{subcaption}
\usepackage{multirow}
\usepackage{wrapfig}
\usepackage{graphicx}
\usepackage{amsmath}
\usepackage{enumerate}
\usepackage{enumitem}

 \usepackage{booktabs}
\usepackage[normalem]{ulem}

\useunder{\uline}{\ul}{}

\newcommand{\mat}[1]{{\bf #1}}

\newtheorem{proposition}{Proposition}[section]
\newtheorem{theorem}{Theorem}[section]

\newtheorem{lemma}[theorem]{Lemma}

\DeclareMathOperator*{\argmax}{arg\,max}

\begin{document}
\title{Fairness through Aleatoric Uncertainty}


\author{Anique Tahir}

\affiliation{%
  \institution{Arizona State University}
  \streetaddress{1151 S Forest Ave}
  \city{Tempe}
  \state{Arizona}
  \country{USA}
}
\email{artahir@asu.edu}

\author{Lu Cheng}
\affiliation{%
  \institution{University of Illinois Chicago}
  \streetaddress{1200 West Harrison St.}
  \city{Chicago}
  \state{Illinois}
  \country{USA}}
\email{lucheng@uic.edu}

\author{Huan Liu}
\affiliation{%
  \institution{Arizona State University}
  \streetaddress{1151 S Forest Ave}
  \city{Tempe}
  \state{Arizona}
  \country{USA}
}
\email{huanliu@asu.edu}

\renewcommand{\shortauthors}{Anique Tahir, Lu Cheng, \& Huan Liu}

\begin{abstract}
\label{sec:abstract}
We propose a simple yet effective solution to tackle the often-competing goals of fairness and utility in classification tasks. While fairness ensures that the model's predictions are unbiased and do not discriminate against any particular group or individual, utility focuses on maximizing the model's predictive performance. 
This work introduces the idea of leveraging aleatoric uncertainty (e.g., data ambiguity) to improve the fairness-utility trade-off. Our central hypothesis is that aleatoric uncertainty is a key factor for algorithmic fairness and samples with low aleatoric uncertainty are modeled more accurately and fairly than those with high aleatoric uncertainty. We then propose a principled model to improve fairness when aleatoric uncertainty is high and improve utility elsewhere. Our approach first intervenes in the data distribution to better decouple aleatoric uncertainty and epistemic uncertainty. It then introduces a fairness-utility bi-objective loss defined based on the estimated aleatoric uncertainty. Our approach is theoretically guaranteed to improve the fairness-utility trade-off. Experimental results on both tabular and image datasets show that the proposed approach outperforms state-of-the-art methods w.r.t. the fairness-utility trade-off and w.r.t. both group and individual fairness metrics. This work presents a fresh perspective on the trade-off between utility and algorithmic fairness and opens a key avenue for the potential of using prediction uncertainty in fair machine learning.

\end{abstract}





\maketitle

\section{Introduction}
\label{sec:introduction}


Machine learning (ML) algorithms have been widely used in various applications and are becoming increasingly popular in domains such as computer vision, speech recognition, natural language processing, and bioinformatics~\cite{lin2022survey}. Despite their superior performance in terms of prediction accuracy, they have often faced criticism for lacking fairness and discriminating against marginalized groups \cite{zemel2013learning,hardt2016equality}. Fair ML aims to improve algorithmic fairness. Due to the often competing relation between fairness and utility, a primary challenge in fair ML has been improving the fairness-utility trade-off \cite{hardt2016equality,mehrabi2021survey}. Finding a solution that alleviates the trade-off and improves both goals is often deemed impossible yet crucial to ensure that ML algorithms are not only functional but also trustworthy when making predictions \cite{chouldechova2017fair,cheng2021socially}.

    

Prior work in fair ML improves training procedures based on certain heuristics (e.g., using an adversary~\cite{zhang2018mitigating}) to achieve a better trade-off~\cite{feldman2015certifying, calmon2017optimized,lahoti2020fairness} (see more works discussed in depth in Section~\ref{sec:related}). In essence, doing so is analogous to finding a better hypothesis to reduce uncertainty in areas where there is a lack of data or knowledge~\cite{liupushing, dutta2020there}. This kind of uncertainty is known as \textit{epistemic} or \textit{model uncertainty} \cite{abdar2021review}. By contrast, this work proposes to explore the connection between fairness and the other kind of predictive uncertainty, known as \textit{aleatoric} \cite{abdar2021review} or \textit{data uncertainty}, arising from the inherent ambiguity in the data.

Aleatoric uncertainty naturally relates to both algorithmic fairness and utility. When data is ambiguous due to e.g., inherent noise or entangled causal features, we humans tend to make decisions relying on past experience and ambiguous information that might reflect historical inequalities. Similarly, ML models are more likely to make wrong predictions under high aleatoric uncertainty, and even if we train on an infinite amount of data, the model would still be uncertain about the prediction~\cite{hullermeier2021aleatoric}. Therefore, \textit{our central hypothesis is that aleatoric uncertainty is a crucial cause of algorithmic unfairness, and samples with low aleatoric uncertainty are modeled more accurately and fairly than those with high aleatoric uncertainty}. The relation between aleatoric uncertainty and fairness has evaded investigation in the past since aleatoric uncertainty is associated with the impossibility of improvement. 

\begin{figure*}[h]
    \centering
    \includegraphics[width=\linewidth]{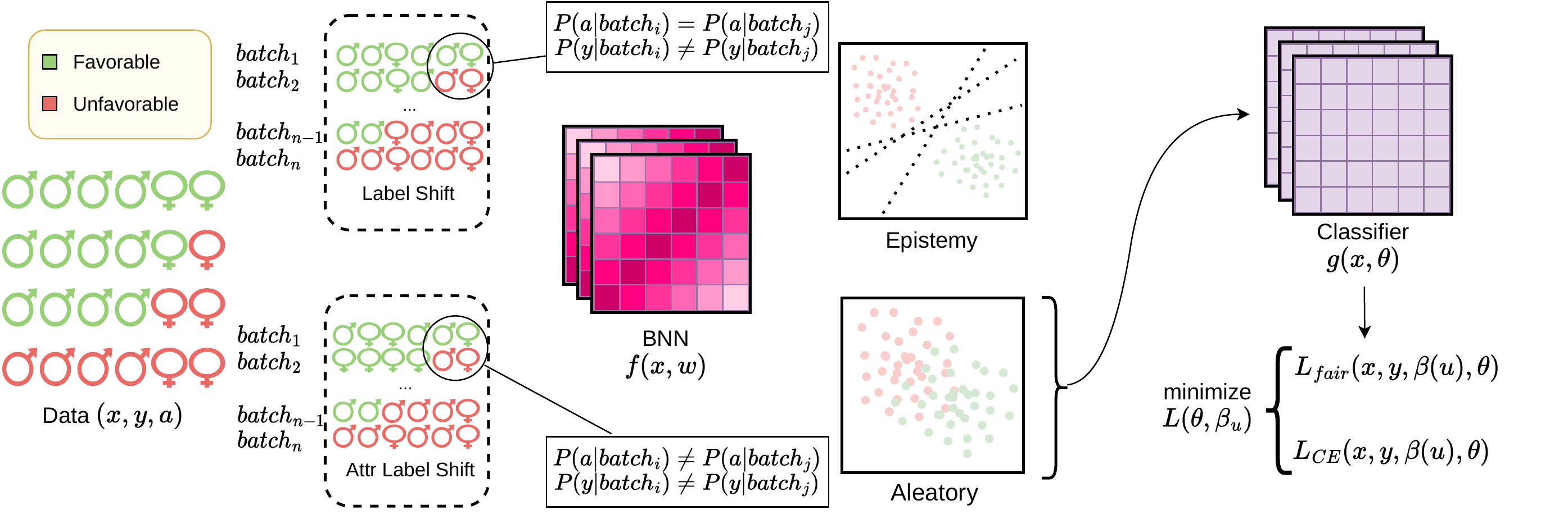}
    \caption{After the distributional intervention, GAIA improves the fairness-utility trade-off by balancing the utility ($L_{CE}$) and fairness ($L_{fair}$) loss using aleatoric uncertainty estimated by BNNs.}
    \label{fig:architecture}
\end{figure*}
To bridge the gap, this work introduces a simple yet effective approach that leverages aleatoric uncertainty to improve the fairness-utility trade-off with theoretical guarantees. In particular, given the potential confounding effects related to the protected attribute, we first propose effective distributional interventions to prevent noise leakage in uncertainty estimation to enable the disentanglement of aleatoric and epistemic uncertainties. 
Predictions with low uncertainty tend to be fair while those with high uncertainty tend to be unfair (Section~\ref{sec:thr_fair_util_tradeoff}). 
Thus, we explicitly model aleatoric uncertainty in the training process: considering heteroscedastic uncertainty (i.e., the uncertainty varies across samples), we prioritize utility over fairness when dealing with samples that have low aleatoric uncertainty, and prioritize fairness over utility for samples with high aleatoric uncertainty. The representation of various protected groups is heterogeneous in real-world data. Conventionally, the ground truth labels provided are assumed to be correct. However, ML models learn spurious correlations since subgroups of the population achieve different distributions of favorable or unfavorable outcomes. This results in algorithmic bias. For our approach, we draw a dichotomy between the solution space; (i) where our model is likely to make the correct prediction, resulting in lower algorithmic bias, and (ii) where it is likely to be uncertain, resulting in higher algorithmic bias. By utilizing this knowledge during model training, we can reduce the trade-off between utility and fairness objectives.
We evaluate our approach on well-established datasets and compare it to the state-of-the-art baselines that include pre-, in-, and post-processing methods \cite{aif360-oct-2018, zemel2013learning}. Experimental results indicate that the proposed approach achieves the best fairness-utility trade-off in terms of both group fairness metrics \cite{dwork2012fairness, zemel2013learning}, and shows potential for individual fairness.

In summary, we introduce several important contributions to the field of fairness in ML:
\begin{enumerate}[label=(\roman*), leftmargin=*]
    \item we provide the first empirical results regarding the relationship among fairness, utility, and aleatoric uncertainty in classification tasks;
    \item we propose a simple yet effective approach that leverages aleatoric uncertainty to improve the fairness-utility trade-off with a theoretical guarantee; and
    \item we provide empirical evidence of its efficacy on real-world datasets. Experimental results also highlight the importance of distributional intervention for uncertainty estimation that would otherwise lead to algorithmic unfairness.
\end{enumerate}

\section{Methodology}
\label{sec:methodology}
\textbf{Problem Setting.} 
We consider the standard fair binary classification setting where the samples $X \in \mathcal{X} \subset \mathbb{R}^n$, labels $Y \in \mathcal{Y} = \{0, 1\}$, and protected attribute $A \in \{0, 1\}$ are provided as the input. Our objective is to train a classifier $g:\mathbb{R}^n \rightarrow [0,1]$ such that its predictions $\hat{Y} \in [0, 1]$ are accurate i.e., $P(\hat{Y}|X) = P(Y|X)$, and fair across different demographic groups. The proposed approach, \textbf{G}uided \textbf{A}lgorithm for \textbf{I}ntegrating \textbf{A}leatory (GAIA), draws from the inherent relation between fairness and aleatoric uncertainty due to data ambiguity which leads a model to rely on biased priors. With high aleatoric uncertainty, it becomes infeasible to improve the utility; however, we can still improve fairness since it does not necessarily rely on utility. We empirically and theoretically prove that GAIA improves the fairness-utility trade-off. GAIA consists of three major steps highlighted in the following subsections.
\raggedbottom
\subsection{Distributional Intervention}
Traditional ML algorithms use Empirical Risk Minimization (ERM)~\cite{donini2018empirical} and rely on the independent and identically distributed (\textit{i.i.d.}) assumption. Prior work shows that distribution shift exacerbates both fairness and predictive performance~\cite{tahir2022distributional,rezaei2021robust}. In addition, due to the skewed distributions for different protected groups, standard uncertainty estimation methods such as BNNs cannot be directly applied to estimating the model uncertainty given its sensitivity to data imbalance~\cite{podkopaev2021distribution}. 
To assuage this issue, we intervene in the data distribution and identify two instances of data bias that can be controlled: the label distribution is skewed resulting in the model relying on (i) the prior distribution of the label (\textit{Label Shift}), and (ii) the spurious correlation between the protected attribute and the label (\textit{Attribute Label Shift}). An example for (i) is when the majority of the data has a specific label (e.g., non-fraud transactions in fraud detection). Here, a trained model may rely on shortcut learning~\cite{geirhos2020shortcut} to predict the majority label. Similarly, for (ii), a trained model may rely on the protected attribute for prediction if it displays a significant correlation with the label. If the protected attribute is correlated with the label, the non-protected covariates affected by the protected attribute also resonate with the correlation. Thus, if we intervene in the correlation between the protected attribute and the label, it also results in the intervention of the factors resonating with the protected attributes in the non-protected covariates.

Distribution intervention can mitigate unfairness and lead to better uncertainty disentanglement. Note that this step can be replaced with other heuristics for achieving better utility and fairness as highlighted in Section.~\ref{sec:related}. This is because using a good heuristic reduces epistemic uncertainty, leading to better estimation of aleatoric uncertainty.\\
\subsubsection{\textbf{Label Shift.}}
\label{sec:lblshft}
Label shift aims to change the distribution of the labels in every mini-batch during training. This will result in a model that does not favor the majority label in the original data distribution. Formally, let $(X, Y) \in \mathcal{D}$ be instances in the dataset $\mathcal{D}$, where $X$ denotes the feature matrix and $Y \in \{0,1\}$ denotes the binary label vector. We define the sets of indices $M_1 = \{i \in \mathcal{D} \mid Y_i = 1\}$ and $M_0 = \{i \in \mathcal{D} \mid Y_i = 0\}$, corresponding to samples with favorable (e.g., low credit risk) and unfavorable outcomes, respectively. $|M_1|= n_1 $ and $|M_0|=n_0$.

A random percentage of favored samples, $p$, is determined by sampling randomly from the uniform distribution $\mathcal{U}(0,1)$. We then define the scaled sets of indices $M_{1}^{'} = \{i \in M_1 \mid p\}$ and $M_{0}^{'} = \{i \in M_0 \mid 1-p\}$. These sets are used to calculate the probability of selecting each sample, $P^1_i = \frac{[i \in M_1^{'}]}{n_1}$, $P^0_i = \frac{[i \in M_0^{'}]}{n_0}$. A batch of size $m$ is selected from the dataset by randomly picking samples without replacement according to the probabilities $P_i$. We denote the set of indices of the selected samples by $I = \{i_1,i_2, \dots, i_m\} \subset \mathcal{D}$. This results in a counterfactual batch of training samples with the intervention of label distribution~\textbf{(LabelShift, LS)}.
\subsubsection{\textbf{Attribute Label Shift.}}
\label{sec:attrlblshft}
Intervening only on the label distribution may be insufficient to reduce the spurious correlations in the data. 
We further intervene on the protected attribute to resolve its confounding effect. However, this is often infeasible with observational data. Therefore, we introduce an estimation of intervention by changing the correlation of protected attribute and label distributions across different mini-batches during training. The underlying assumption is that there is a sufficient number of non-causal factors in the covariates such that the interventional changes are large enough for the model to distinguish the non-causal factors from the causal ones. Thus, Attribute Label Shift aims to intervene on both the protected attribute $a$ and the label $y$. Let $M_{p^1} = \{i \in \mathcal{D} \mid a_i = 1\}$ and $M_{p^0} = \{i \in \mathcal{D} \mid a_i = 0\}$ be the sets of indices for samples belonging to the protected group and non-protected group, respectively. $n_{p^{1}} = |M_{p^{1}}|$ and $n_{{p^{0}}} = |M_{{p^{0}}}|$.

A random percentage $p_1$ of samples from the protected group is determined by sampling randomly from the uniform distribution $\mathcal{U}(0,1)$. We then define the scaled sets of indices $M'_{p^{1}} = \{i \in M_{p^{1}} \mid p_1\}$ and $M'_{p^{0}} = \{i \in M_{p^{0}} \mid 1-p_1\}$. These sets are used to calculate the probability of selecting each sample $P_{p,i}$ from the protected or non-protected group, $P_{p^1,i} = \frac{[i \in M'_{p^{1}}]}{n_{p^{1}}}$, $P_{p^0,i} = \frac{[i \in M'_{p^{0}}]}{n_{{p^{0}}}}$.
Similarly, the probability of selecting each sample $P_{f,i}$ from the favored or unfavored class is defined as $P_{1,i} = \frac{[i \in M_1^{'}]}{n_1}$, $P_{0,i} = \frac{[i \in M_0^{'}]}{n_0}$. The final probability of selecting each sample is the product, $P_i = P_{p,i} * P_{f,i}$. This gives us a batch of training samples with interventions on the correlation of protected attribute and label~\textbf{(AttrLabelShift, ALS)}.

\subsection{Decoupling Aleatoric and Epistemic Uncertainty}
GAIA uses BNN via backpropagation (Bayes by Backprop)~\cite{blundell2015weight} to conveniently decouple aleatoric from epistemic uncertainty while also maintaining its ability to be integrated into existing neural architectures. Bayes by Backprop is computationally efficient and theoretically sound. Given $C$ classes, aleatoric uncertainty is formulated as the expected entropy for the prediction \cite{abdar2021review, hullermeier2021aleatoric}, 
\begin{equation}
    H_{\text{alea}}(\mat{x}) = \int_\theta \sum_i^C -p(y_i|\mat{x},\theta)\log p(y_i|\mat{x},\theta) \,d\theta,
    \label{eq:alea}
\end{equation}
where $p(y_i|x,\theta)$ is the predictive probability of the $i$-th class from the model parameterized by $\theta$. Epistemic uncertainty is represented by the model's predictive variance \cite{abdar2021review}, 
\begin{equation}
    \sigma^2_{\text{epi}}(\mat{x}) = \text{Var}_j[p(y|\mat{x},\theta_j)],
    \label{eq:epistemic}
\end{equation}
where $j$ denotes the $j$-th sample of the BNN weights. 
BNN involves finding the maximum a posteriori~(MAP) weights:
\begin{equation}
        \theta^{MAP} = \argmax_\theta \log P(\theta|\mathcal{D}) 
                     = \argmax_\theta P(\mathcal{D}|\theta) + \log P(\theta).
\end{equation}
The final prediction of BNNs is the expected value of the predicted label $\hat{y}$ for an unseen sample $\hat{x}$ over the posterior distribution of the weights, $P(\theta|\mathcal{D})$ i.e., $P(\hat{y}|\hat{x}) = \mathbb{E}_{P(\theta|\mathcal{D})} [P(\hat{y}|\hat{x}, \theta)]$. We can then utilize each candidate prediction, $P(\hat{y}|\hat{x}, \theta_j)$, where $\theta_j \sim P(\theta|\mathcal{D})$ to efficiently evaluate both aleatoric and epistemic uncertainties using Eq.~\ref{eq:alea} and \ref{eq:epistemic}, respectively. 

For tractable estimation, the common practice in variational inference estimates the posterior using a surrogate, $q(\theta|w)$, by minimizing the Evidence Lower Bound (ELBO) loss~\cite{yang2017understanding}. Further, we assume heteroscedastic uncertainty, i.e., uncertainty varies across different samples \cite{cao2020heteroskedastic}, given its practicability. Hence, the uncertainty metrics between predictions are on a per-sample basis. By explicitly modeling aleatoric and epistemic uncertainty, GAIA traces whether the uncertainty stems from ambiguity or lack of data.

\subsection{Improving Fairness-Utility Trade-off}
The goal of GAIA is leveraging aleatoric uncertainty to bridge the gap between fairness and accuracy based on the hypothesis that samples with low aleatoric uncertainty are modeled more accurately and fairly than those with high uncertainty. Thus, to achieve a better trade-off, we design a model to improve fairness when aleatoric uncertainty is high and improve utility elsewhere. 
We first describe the function $\beta(u): \mathbb{R}^m \rightarrow \mathbb{R}^m$ that assigns weights to samples based on the estimated aleatoric uncertainty $u$:
\begin{equation}
    \beta(u) = \left(\frac{u - u_{min}}{u_{max} - u_{min}}\right)^k,
\end{equation}
where the hyper-parameter $k$ helps to weigh one objective in favor of the other, and $u_{min}$ and $u_{max}$ are two hyperparameters to normalize the weights.
The overall objective function of GAIA (Eq.~\ref{eq:final}) is a bi-objective loss corresponding to both utility and fairness. It maximizes utility for the samples with low aleatoric uncertainty; for samples with high aleatoric uncertainty, there is little improvement to be made in terms of utility due to the inherent ambiguity of the data. Thus, the aim of GAIA is to steer the objective toward improving the fairness of samples with high aleatoric uncertainty.

Given a batch of training data $\mathcal{S} \subseteq \mathcal{D}$ and a classifier parameterized by $\theta$, the utility loss is a  weighted cross-entropy loss:
\begin{equation} \label{eq:weighted_ce}
\begin{split}
    \mathcal{L}_{CE}(\mathcal{S}, \beta) = -\frac{1}{|\mathcal{S}|}\sum_{i \in \mathcal{S}} &\beta_i\big(y_i\log(p(y_i|x_i, u_i))+ \\ & (1-y_i)\log(1-p(y_i|x_i, u_i))\big),
\end{split}
\end{equation}
where $y_i$ is the label for sample $x_i$ and $\beta_i = \beta(u_i)$. The conditioning on prediction $p(y_i|x_i, u_i)$, allows the model to make an informed choice based on the uncertainty. We define fairness as the difference in the mean cross-entropy between instances of different protected attributes. We show in Section~\ref{sec:Theory} that our proposed metric acts as a feasible surrogate to cover common group fairness metrics. Let $\mathcal{S}_0$ and $\mathcal{S}_1$ be the sets of samples whose protected attribute is $0$ and $1$, respectively. We define fairness as follows:
\begin{equation} \label{eq:fair}
\mathcal{L}_{fair}(\mathcal{S}, 1-\beta) = |\mathcal{L}_{CE}(\mathcal{S}_0, 1-\beta) - \mathcal{L}_{CE}(\mathcal{S}_1, 1-\beta)| \quad \mathcal{S}_0\cup \mathcal{S}_1=\mathcal{S}.
\end{equation}
The objective function of GAIA, $\mathcal{L}$, is the sum of Eq.~\ref{eq:weighted_ce} and Eq.~\ref{eq:fair}:
\begin{equation} \label{eq:final}
    \mathcal{L}(\mathcal{S},\beta) = \mathcal{L}_{CE}(\mathcal{S}, \beta) + \mathcal{L}_{fair}(\mathcal{S}, 1-\beta).
\end{equation}

\section{Theoretical Guarantee to Improve the Trade-off}
\label{sec:Theory}
In this section, we theoretically prove GAIA can guarantee to improve the fairness-accuracy trade-off through the following three key hypotheses: (i) as aleatoric uncertainty increases, accuracy will decrease; (ii) we can improve fairness in regions of high aleatoric uncertainty; and (iii) binary cross-entropy (BCE) \textit{difference} across separate protected groups (Eq. \ref{eq:fair}) is proportional to common group fairness metrics such as equal opportunity difference (EOD) and average odds difference (AOD). The proof consists of two propositions. First, we show divergence on the optimal utility under aleatoric uncertainty. Second, we show the convergence for fairness under BCE difference between protected and non-protected groups. As per convention from the problem setting and for the sake of simplicity, we consider the binary classification case. We use AOD for illustration and similar formulation extends to other group fairness metrics such as EOD.
\subsection{Relation between Aleatoric Uncertainty and Accuracy}
\begin{theorem}
\label{thm:acc_alea_text}
As the aleatoric uncertainty increases, the model's accuracy approaches random chance:
\[
\lim_{\mathbb{E}[H[q(y|x)]] \rightarrow \inf} \text{accuracy} = \frac{1}{C},
\]
\end{theorem}
\noindent where $C$ is the number of classes. 

\noindent\textit{Proof of Theorem \ref{thm:acc_alea_text}.} We first define the predictive entropy for the model. Let $p(y|x)$ be the predicted probability distribution of the target class $y$ given the input instance $x$. In a binary classification problem where $y \in \{0, 1\}$, the expected predictive entropy is the average predictive entropy over all instances in the dataset. This represents the aleatoric uncertainty (Eq.~\ref{eq:alea}).

Next, we will show that the lower bound on the accuracy approaches random chance as the expected predictive entropy increases. In binary classification, random chance corresponds to an accuracy of \nicefrac{1}{2}, suggesting that the model is not better than random guessing. We first derive a lower bound on the accuracy using Fano's inequality~\cite{scarlett2019introductory}. Fano's inequality relates the conditional probability of error in predicting the target class $y$ given the input instance $x$ with the mutual information between $y$ and $x$:
\begin{lemma}[Fano's inquality]
\[ H(\epsilon) + \epsilon \log(C - 1) \geq H(Y|X) ,\]
\end{lemma}
\noindent where $H(\epsilon)$ is the binary entropy function of $\epsilon$, the probability of error in predicting the target class, and $H(Y|X)$ is the conditional entropy of the true conditional probability distribution. In a binary classification problem, $C=2$ and we can simplify Fano's inequality as follows:
\begin{equation}
H(\epsilon) + \epsilon \log(1) \geq H(Y|X) .
\end{equation}
Since $\log(1) = 0$, the inequality becomes:
\begin{equation}
H(\epsilon) \geq H(Y|X).
\end{equation}
The probability of error $\epsilon$ is related to the accuracy by the following relationship:
\begin{equation}
\epsilon = 1 - \text{Accuracy}.
\end{equation}
\noindent We can then reformulate Fano's inequality in terms of accuracy:
\begin{equation}
H(1 - \text{Accuracy}) \geq H(Y|X).
\end{equation}
Since the binary entropy function $H(p)$ is a monotonically increasing function for $0 \leq p \leq \nicefrac{1}{2}$ and a monotonically decreasing function for $\nicefrac{1}{2} \leq p \leq 1$, the maximum entropy is achieved when $p = \nicefrac{1}{2}$. Thus, the entropy of the error probability is maximized when the accuracy is at random chance:
\begin{equation}
H(1 - \nicefrac{1}{2}) = H(\nicefrac{1}{2}) = 1.
\end{equation}
Therefore, as the expected predictive entropy $\mathbb{E}[H[q(y|x)]]$ increases, the lower bound on the accuracy given by Fano's inequality approaches the maximum entropy state, which corresponds to random chance.

\subsection{Relation between BCE Loss Difference and Fairness}
\begin{theorem}
\label{thm:bce_diff_fair_text}
The expected difference in BCE losses between the protected and non-protected groups defined in Eq. \ref{eq:fair} is proportional to the Average Odds Difference (AOD).
\[ \mathbb{E}[\Delta L(y)] = \frac{1}{N_1}\sum_{i \in A_1} \Delta L(y_i) - \frac{1}{N_2}\sum_{j \in A_2} \Delta L(y_j) \propto \text{AOD}. \]
\end{theorem}
\noindent\textit{Proof of Theorem \ref{thm:bce_diff_fair_text}.} Let us denote the protected attribute instances as $A_1$ and $A_2$. Let $p_i$ be the predicted probability of the positive class $y = 1$ for instances in the group with protected attribute $A_i$, where $i \in \{1, 2\}$.

\begin{proposition}
The Binary Cross-Entropy (BCE) loss for instances with protected attribute $A_i$ is given by
\[ L_i(y, p_i) = - y \log(p_i) - (1 - y) \log(1 - p_i). \]
\end{proposition}

This proposition follows directly from the definition of BCE for binary classification problems. For group fairness metrics, we are concerned with True Positive Rate~($\text{TPR}_i$) difference and False Positive Rate~($\text{FPR}_i$) difference between  different groups. 

\begin{lemma}[Average Odds Difference]
\label{lem:aod}
The Average Odds Difference (AOD) between group $A_1$ and group $A_2$ is given by
\[ \text{AOD} = \frac{\big|\text{TPR}_1 - \text{TPR}_2\big| + \big|\text{FPR}_1 - \text{FPR}_2\big|}{2}. \]
\end{lemma}

Now, let us analyze the difference between the BCE losses for the protected ($L_1(\cdot)$) and non-protected ($L_2(\cdot)$)groups:

\begin{lemma}
\label{lem:bce_prot}
The difference in BCE losses between the two protected attribute groups $A_1$ and $A_2$ can be expressed as
\begin{equation*}
\begin{split}
\Delta L(y) &= L_1(y, p_1) - L_2(y, p_2) \\
             &= - y \log\left(\frac{p_1}{p_2}\right) - (1 - y) \log\left(\frac{1 - p_1}{1 - p_2}\right) .
\end{split}
\end{equation*}

\end{lemma}

Let $N_1$ and $N_2$ be the total number of instances in the protected $A_1$ and non-protected groups $A_2$, respectively. To prove Theorem~\ref{thm:bce_diff_fair_text}, we compute the expected differences in BCE losses for the true positive and false positive cases separately.
\subsubsection{\textbf{Equal Opportunity Difference and BCE Difference}}
First, consider the true positive cases where $y = 1$. In this case, $\Delta L(y=1) = - \log(\frac{p_1}{p_2})$ (from Lemma~\ref{lem:bce_prot}). The expected difference in BCE losses for true positives in both groups can be expressed as:
\begin{equation}
\label{eq:bced_eod}
\begin{split}
\mathbb{E}[\Delta L(y = 1)] &= \frac{1}{N_1}\sum_{i \in A_1, y_i = 1} -\log\left(\frac{p_1}{p_2}\right) - \frac{1}{N_2}\sum_{j \in A_2, y_j = 1} -\log\left(\frac{p_1}{p_2}\right) \\
                        &\propto \big|\text{TPR}_1 - \text{TPR}_2\big| = EOD .
\end{split}
\end{equation}

\subsubsection{\textbf{Average Odds Difference and BCE Difference}}
Next, consider the false positive cases where $y = 0$. In this case, $\Delta L(y) = - \log(\frac{1 - p_1}{1 - p_2})$. The expected difference in BCE losses for false positives in both groups can be expressed as:
\begin{equation}
\label{eq:bced_fprd}
\begin{split}
\mathbb{E}[\Delta L(y = 0)] &= \frac{1}{N_1}\sum_{i \in A_1, y_i = 0} -\log\left(\frac{1 - p_1}{1 - p_2}\right) \\
& - \frac{1}{N_2}\sum_{j \in A_2, y_j = 0} -\log\left(\frac{1 - p_1}{1 - p_2}\right) \\
&\propto \big|\text{FPR}_1 - \text{FPR}_2\big| .
\end{split}
\end{equation}
Finally, by combining the expected differences in BCE losses for true positive~(Eq.~\ref{eq:bced_eod}) and false positive~(Eq.~\ref{eq:bced_fprd}) cases with Lemma~\ref{lem:aod}, we get:
\begin{equation}
\begin{split}
\mathbb{E}[\Delta L(y)] &= \mathbb{E}[\Delta L(y = 1)] + \mathbb{E}[\Delta L(y = 0)] \\
&\propto \big|\text{TPR}_1 - \text{TPR}_2\big| + \big|\text{FPR}_1 - \text{FPR}_2\big| = \text{AOD} \times 2 .
\end{split}
\label{eq: proportion}
\end{equation}

Thus, Eq. \ref{eq: proportion} shows that the expected difference in BCE losses between the two protected attribute groups is proportional to AOD. This implies that minimizing the difference in BCE losses can lead to fairer outcomes with respect to AOD. EOD is a subset of AOD as demonstrated by Eq.~\ref{eq:bced_eod}.

\subsubsection{\textbf{A Closer Look}}
Here, we elaborate on why Eq. \ref{eq:bced_eod} and Eq. \ref{eq: proportion} hold. From Lemma \ref{lem:bce_prot}, for the true positive cases where $y = 1$, we have $\Delta L(y = 1) = - \log\left(\frac{p_1}{p_2}\right)$. We first analyze the relationship between the expected difference in BCE losses and the TPR for the two protected attribute groups.

Denote the total number of true positive instances for each group as $N_1^{TP}$ and $N_2^{TP}$, and let $\text{TPR}_1$ and $\text{TPR}_2$ be the true positive rates for the groups $A_1$ and $A_2$, respectively. The expected difference in BCE losses for the true positive instances is represented as
\begin{equation}
\label{eq:s_1}
\begin{split}
\mathbb{E}[\Delta L(y = 1)] &= \frac{1}{N_1^{TP}}\sum_{i \in A_1, y_i = 1} -\log\left(\frac{p_1}{p_2}\right) 
  \\&- \frac{1}{N_2^{TP}}\sum_{j \in A_2, y_j = 1} -\log\left(\frac{p_1}{p_2}\right).
\end{split}
\end{equation}

We reformulate Eq.~\ref{eq:s_1} using TPR values as follows: 
\begin{equation}
\label{eq:s_2}
\begin{split}
\mathbb{E}[\Delta L(y = 1)]&= \frac{1}{\text{TPR}_1 N_1} \sum_{i \in A_1, y_i = 1} -\log\left(\frac{p_1}{p_2}\right) 
\\&- \frac{1}{\text{TPR}_2 N_2} \sum_{j \in A_2, y_j = 1} -\log\left(\frac{p_1}{p_2}\right).
\end{split}
\end{equation}

Eq.~\ref{eq:s_2} indicates that as the difference between $\text{TPR}_1$ and $\text{TPR}_2$ increases, $\mathbb{E}[\Delta L(y = 1)]$ also increases. This means that if there is a notable difference in the TPR between the two groups, it will result in a substantial dissimilarity in the BCE losses as well. Therefore, we can conclude that the expected difference in BCE losses for the true positive cases, $\mathbb{E}[\Delta L(y = 1)]$, is indeed proportional to the difference in TPR between the two protected attribute groups. Similarly, we can establish the proportionality of the expected difference in BCE losses for false positive cases, $\mathbb{E}[\Delta L(y = 0)]$, to the difference in FPR between the groups. Combining the results for true positive and false positive cases, we demonstrate that the expected difference in BCE losses between the two protected attribute groups is proportional to the AOD, as stated in Theorem \ref{thm:bce_diff_fair_text}. In other words, the expected difference in BCE losses for true positive cases captures the difference in TPR and FPR between the two protected attribute groups, which is an essential component of common group fairness metrics such as EOD and AOD.
\subsection{On the Fairness-Utility Trade-off}
\label{sec:thr_fair_util_tradeoff}
Under Theorem~\ref{thm:bce_diff_fair_text}, we show that by minimizing the BCE loss difference in regions of high aleatoric uncertainty, we indirectly improve group fairness, as reducing the loss entails minimizing the disparities across different groups.
In these regions, the model's predictions are more susceptible to biases and disparities since it relies on learned priors, leading to unfair predictions. By prioritizing fairness in these regions, we aim to mitigate the adverse effects of aleatoric uncertainty on marginalized groups. As per Theorem~\ref{thm:acc_alea_text}, it is not feasible to improve accuracy in such regions. 

For regions of high confidence (i.e., low uncertainty), accuracy converges to $1$ (due to the law of large numbers). Thus, when the uncertainty is low, fairness improves. We can achieve fairness by optimizing utility. Based on Lemma~\ref{lem:aod}, we have
\begin{equation}
    \begin{split}
    \lim_{accuracy \rightarrow 1} AOD &= \dfrac{\big|\text{TPR}_1 - \text{TPR}_2\big| + \big|\text{FPR}_1 - \text{FPR}_2\big|}{2}\\
    &=\dfrac{|1-1| + |0-0|}{2} = 0.
    \end{split}
\end{equation}

According to Theorems \ref{thm:acc_alea_text} and \ref{thm:bce_diff_fair_text}, GAIA targets utility and fairness in the respective regions where the other metric is non-conflicting. This results in the improvement of both utility and fairness while minimizing the trade-off.

\section{Experiments}
\label{sec:experiments}
\begin{figure*}
    \centering
    \includegraphics[width=.8\textwidth]{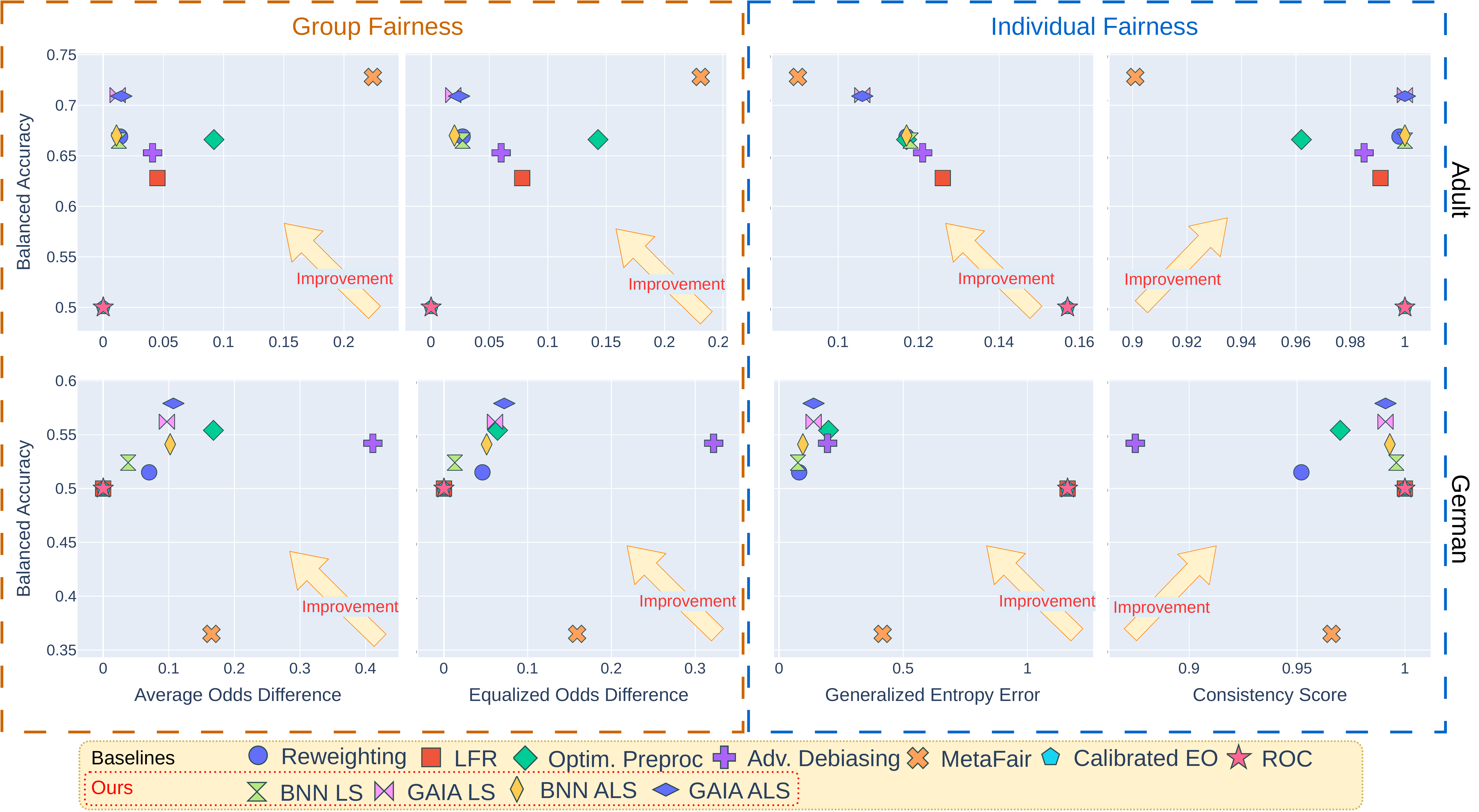}
    
    \caption{Comparison of \textit{Group} (left) and \textit{Individual} (right) Fairness for the \textit{Adult} and \textit{German} Datasets. Various approaches fall on different places on the Pareto front representing the fairness-utility trade-off. }
    \label{fig:dataset_method_comparison}
\end{figure*}
In this section, we show empirical evidence of the effectiveness of GAIA. We aim to answer the following research questions: 
\begin{itemize}[leftmargin=*]
    \item \textbf{RQ1:} How does GAIA fare against the state-of-the-art baselines in terms of the fairness-utility trade-off? 
    \item \textbf{RQ2:} How does empirical evidence support our hypothesis regarding aleatoric uncertainty,  fairness, and utility?
    \item \textbf{RQ3:} While designed for group fairness, what role does GAIA play in improving individual fairness?
\end{itemize}

\subsection{Experimental Setup}
Experiments are conducted for both tabular and image datasets. For tabular data, we compare GAIA with seven baselines including common pre-processing, in-processing, and post-processing approaches. We use two benchmark tabular datasets and four fairness metrics including both group and individual fairness metrics. In particular, for \textbf{RQ. 1-2}, we use EOD and AOD as the group fairness metrics. We use Generalized Entropy Error (GE)~\cite{speicher2018unified} and Consistency Score (CS)~\cite{zemel2013learning} to measure individual fairness for \textbf{RQ. 3}. For utility measure, we use balanced accuracy, which is conventionally used in fairness literature since it captures balanced protected groups. For the image classification task, we use one benchmark dataset and two additional state-of-the-art approaches as baselines to validate the generalizability of GAIA. 

\noindent\textbf{Datasets.} 
The benchmark tabular datasets and image dataset for fair machine learning are detailed below:
\begin{itemize}[leftmargin=*]
    \item \textbf{Adult~\cite{misc_adult_2}:} This dataset consists of multiple features ranging from work class, age, education, and sex. Each instance has a binary label based on whether an individual's income exceeds \$50,000/yr. This dataset consists of 48,842 samples.
    \item \textbf{German~\cite{german_dataset}:} This dataset consists of features related to the financial status of individuals. The label represents whether the attributes represent good or bad credit risk. This dataset consists of 1,000 samples.
    \item \textbf{CelebA~\cite{liu2015faceattributes}:} This dataset contains aligned faces of celebrities with annotations of various attributes, such as gender, age, expression, hair type, and attractiveness. This dataset contains 202,599 face images from 10,177 celebrities.
\end{itemize}
Gender is considered as the protected attribute in each dataset. Features in tabular datasets are binarized, preprocessed, and scaled following \textcite{aif360-oct-2018}. Preprocessing for CelebA follows the conventions established by \textcite{chuangfair}.

\noindent\textbf{Baselines.} For tabular data, we compare GAIA against seven well-established baseline approaches. These approaches can be divided into pre-processing, in-processing, and post-processing methods.
\begin{itemize}[leftmargin=*]
    \item \textbf{Reweighting~\cite{kamiran2012data}:} Reweighing is a \textit{pre-processing} approach that adjusts the weight assigned to examples in each (group, label) pairing to promote fairness prior to classification.
    \item \textbf{Learning Fair Representations (LFR)~\cite{zemel2013learning}:} A \textit{preprocessing} technique aimed at discovering a latent representation that effectively encodes the data while concealing information pertaining to protected attributes.
    \item \textbf{Optimized Preprocessing~\cite{calmon2017optimized}:} Optimized preprocessing is a \textit{pre-processing} approach that employs a probabilistic transformation to modify both features and labels in the data while considering fairness with respect to groups, minimizing individual distortion, and preserving data integrity.
    \item \textbf{Adversarial Debiasing~\cite{zhang2018mitigating}:} Adversarial debiasing is an \textit{in-processing} technique that trains a classifier to achieve high prediction accuracy while simultaneously reducing the adversary's capacity to infer protected attributes from the predictions. This results in a fair classifier, as the predictions are rendered devoid of any group discrimination information that could be leveraged by the adversary.
    \item \textbf{MetaFair~\cite{celis2019classification}:} An \textit{in-processing} meta-algorithm for fair classification that handles a broad range of fairness constraints, including non-convex linear fractional constraints such as predictive parity.
    \item \textbf{Calibrated Equalized-Odds~\cite{pleiss2017fairness}:} A \textit{post-processing} technique which uses the calibrated predicted scores to adjust the labels towards better equalized-odds.
    \item \textbf{Reject Option Classification (ROC)~\cite{kamiran2012decision}:} A \textit{post-processing} technique that balances favorable outcomes between privileged and unprivileged groups by altering the decision boundary in regions of the highest uncertainty.
\end{itemize}
To further examine the effectiveness of the incorporated aleatoric uncertainty, we compare GAIA against its two sub-module variants:
\textbf{BNN LS} is the uncertainty estimation component where a BNN is trained using Label Shift (Section~\ref{sec:lblshft}), and \textbf{BNN ALS} where it is trained using Attribute Label Shift (Section~\ref{sec:attrlblshft}). 

The baseline methods for tabular data are not designed for image modality. Thus, for fair comparisons, we consider the following two state-of-the-art approaches for fair image classification:
\begin{itemize}[leftmargin=*]
    \item \textbf{FairBatch:~\cite{rohfairbatch}} FairBatch seeks to improve the batch selection process through bi-level optimization such that the downstream model achieves improved fairness.
    \item \textbf{FairMixup~\cite{chuangfair}:} FairMixup uses data augmentation to improve the fairness-utility tradeoff by making the underlying model more generalizable through regularization on interpolates.
\end{itemize}

 \noindent\textbf{Implementation Details.} 
For the sake of simplicity in our experiments, we employ a logistic regression model, which is essentially a multi-layer perceptron (MLP) without any hidden layers. The uncertainties utilized for training the classification model are generated using a BNN that consists of three hidden layers. The activation functions employed for the BNN and MLP are LeakyReLU~\cite{maas2013rectifier} and ReLU~\cite{agarap2018deep}, respectively. When necessary, we utilize the Adam optimizer~\cite{kingma2014adam}. Both the BNN and MLP are designed using the JAX framework~\cite{jax2018github} and Oryx~\cite{oryx} for sampling from distributions. For image classification, ResNet-18~\cite{he2016deep} is used as the backbone for both the BNN and the final classifier. We provide the source code for our implementation\footnote{https://github.com/aniquetahir/GAIA}.

To select the best model from training, we use a simple approach: During the training phase, between each mini-batch, we calculate the smoothed training prediction accuracy by using a running average. We select the model parameters corresponding to the best-smoothed accuracy during training for inference. For the baselines, we use standard implementations provided by the AI Fairness 360 Toolkit~\cite{aif360-oct-2018} using the recommended hyper-parameters where needed. For image baselines, we follow the open-source code provided by the authors, respectively~\cite{rohfairbatch, chuangfair}. 

\subsection{Experimental Results}
\textbf{Tabular Data.} We present the experimental results for \textbf{RQ1} regarding the trade-off between models' utility and fairness. We visualize the comparison of Pareto fronts regarding group fairness in Fig.~\ref{fig:dataset_method_comparison} (left). Our model displays pareto dominance in most of the cases overall. We observe that the in-processing approaches (Adversarial Debiasing, MetaFair) prefer fairness over utility. In contrast, pre- (Reweighting, LFR, Optimized Preprocessing) and post-processing (Calibrated EO, ROC) approaches have a more balanced trade-off. We also observe a difference in the trade-off across the Adult and German datasets due to variations in their sample sizes. The Adult dataset ($\sim$48k samples) is significantly larger compared to the German dataset (1,000 samples). This may cause each method to perform distinctly from the perspective of the fairness-utility trade-off. 

For the Adult dataset, we see a smaller disparity between the performance for versions of our approach using Label Shift (LS) and Attribute Label Shift (ALS). We hypothesize that this is due to the larger size of the Adult dataset compared to the German dataset. The larger dataset size allows the model to make better generalizations and reduce the uncertainty overall. Thus, the shift used in the BNN training is less relevant. By comparison, we see a more diverse performance for the German dataset. The ALS counterparts of both BNN and GAIA outperform LS in terms of utility. However, we see slightly better fairness from the LS counterparts. We believe this is due to the LS versions falling closer towards random chance which increases fairness since instances of the protected attribute are treated equally random. For GAIA LS and GAIA ALS, the disparity between fairness is less pronounced since both versions perform comparatively better than random chance.

\begin{table}[]
\centering
\begin{tabular}{@{}llll@{}}
\toprule
\textbf{} & FairBatch           & FairMixup              & \textbf{GAIA}          \\ \cmidrule(l){2-4} 
Bal Acc   $\uparrow$       & {\ul 0.562 (0.138)} & 0.549 (0.035)          & \textbf{0.602 (0.065)} \\
AOD        $\downarrow$     & {\ul 0.047 (0.105)} & \textbf{0.041 (0.032)} & 0.108 (0.068)          \\
EOD        $\downarrow$     & {\ul 0.035 (0.077)} & 0.044 (0.040)          & \textbf{0.021 (0.018)} \\
GE        $\downarrow$     & {\ul 0.086 (0.070)} & 0.260 (0.144)          & \textbf{0.079 (0.022)} \\ \bottomrule
\end{tabular}%
\caption{GAIA shows an overall improvement over baselines w.r.t. balanced accuracy, group (AOD and EOD), and individual fairness (GE) metrics on CelebA image dataset.}
\label{tab:gaia_images}
\end{table}

Fig.~\ref{fig:dataset_method_comparison} also illustrates the value of uncertainty-guided training in GAIA which considers a weighted sum of utility and fairness objectives. Even though BNN with distribution shift (BNN LS and BNN ALS) by itself shows competitive performance compared to the baselines, GAIA consistently outperforms the BNN in terms of utility while matching it in terms of fairness. This improvement is more pronounced in the Adult dataset, where there are more samples for GAIA to leverage the disparity between ambiguous and non-ambiguous subsets of data. Our results highlight the viability of GAIA in improving the fairness-utility trade-off (\textbf{RQ1}).
\\
\textbf{Image Data.} To analyze the generalizability of our approach, we also evaluate its performance in the image domain using the Celebrity Faces dataset (CelebA)~\cite{liu2015faceattributes}. We do not report the Consistency Score for fair image classification since the consistency distance in image data at a pixel level is affected by spurious features, such as the background. We highlight our results in Table~\ref{tab:gaia_images}. 

For multi-objective optimization, an outcome is considered Pareto dominant if both utility and fairness are improved~\cite{bernheim1987coalition}. GAIA is Pareto dominant over FairMixup and FairBatch for all compared fairness metrics except for AOD. FairBatch is Pareto dominant in the same metrics over FairMixup. While FairMixup is not Pareto dominant for AOD since it has lower accuracy, it shows superior AOD performance. We hypothesize this is due to its predictions being closer to random chance since random predictions are considered fair under AOD. 

FairBatch uses meta-optimization of the batch selection process to make the underlying model training to be fair. GAIA uses a similar idea for batch selection using Label Shift~(LS), and Attribute Label Shift~(ALS). However, while our approach explicitly intervenes in the label distribution and the attribute-label correlation, FairBatch uses an outer loss that attempts to train the model in batch selection. In addition, GAIA is capable of premonition regarding uncertainty, allowing it to make informed predictions that lead to a better trade-off. In contrast, Fair-Mixup uses data augmentation. The counterfactuals generated by data augmentation through interpolation may not reflect reality. However, when the batch selection process is changed in FairBatch and GAIA, each sample comes from the training data. Thus, while the data distribution changes, each sample reflects a real sample. This explains the superior performance of both FairBatch and GAIA over FairMixup. 

\begin{figure}
    \centering
         \begin{subfigure}[b]{0.45\columnwidth}
         \centering
         \includegraphics[width=\textwidth]{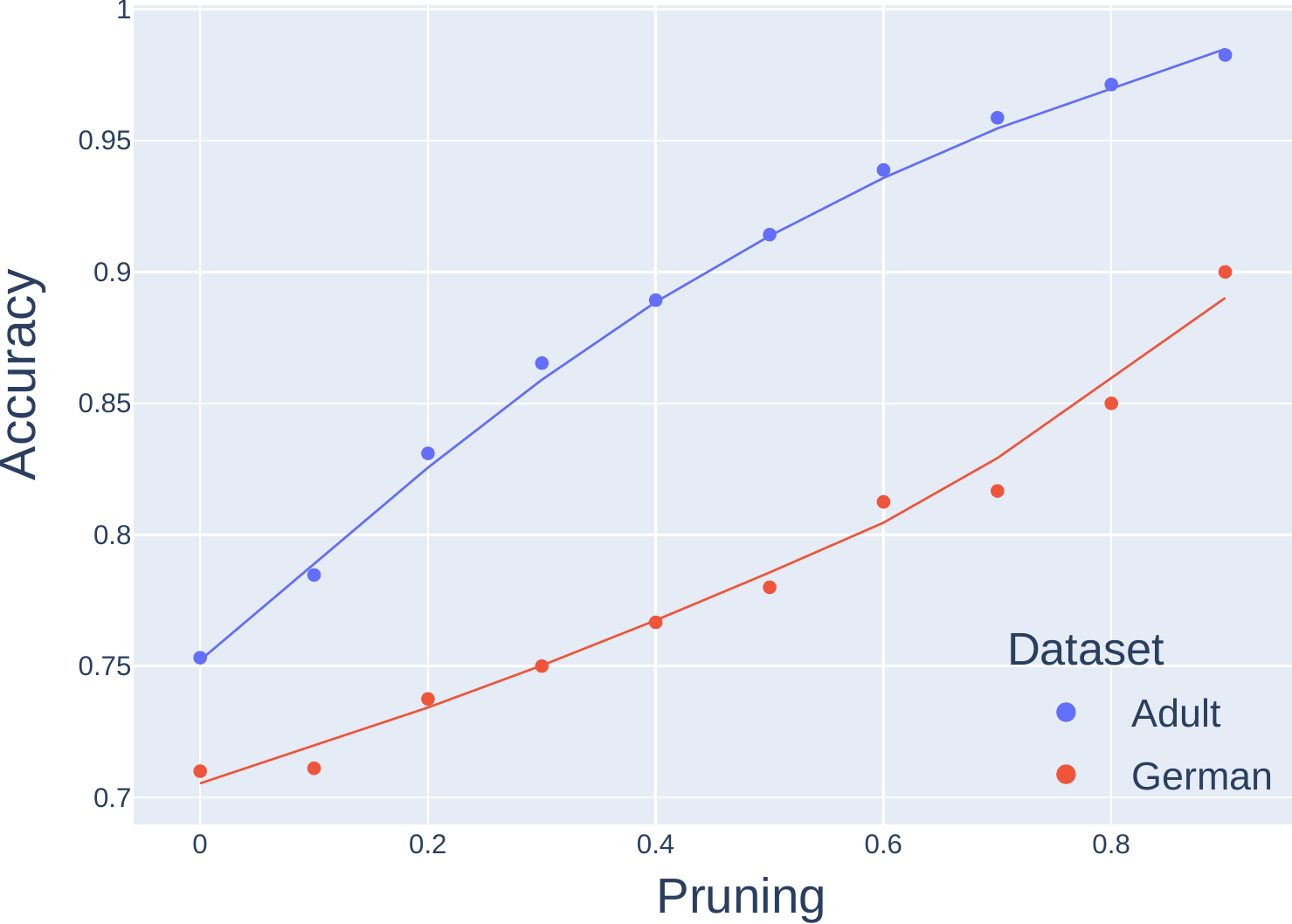}
         \caption{Accuracy}
         \label{fig:suba}
     \end{subfigure}
     \begin{subfigure}[b]{0.45\columnwidth}
         \centering
         \includegraphics[width=\textwidth]{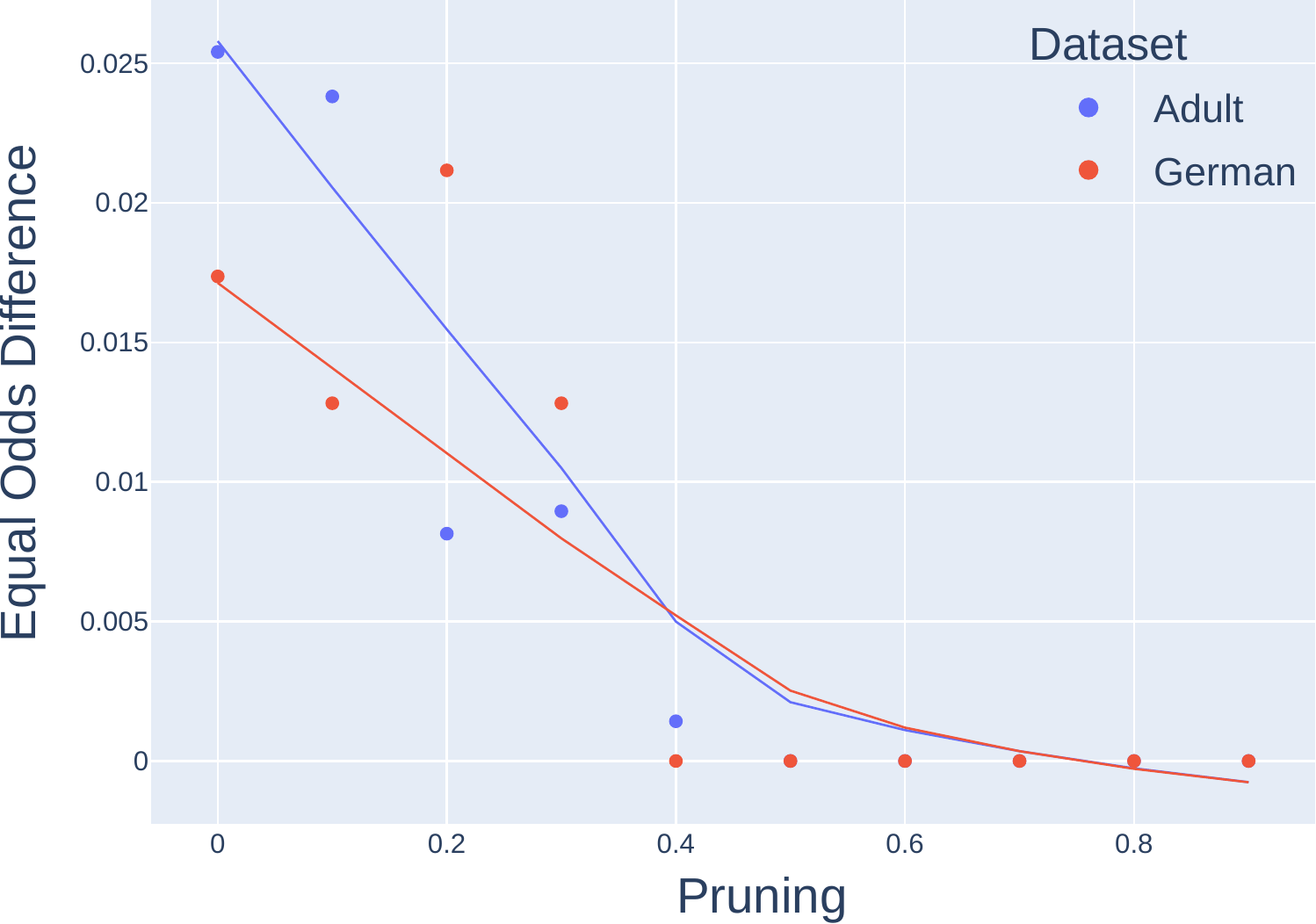}
        \caption{Equal Odds Difference}
        \label{fig:subb}
     \end{subfigure}
    \caption{Pruning the most uncertain samples leads to an improvement in both utility and fairness for the Adult and German datasets. We make similar findings across datasets and various versions of our approach. }
    \label{fig:pruning}
\end{figure}

\subsection{Relation among Aleatoric Uncertainty, Utility, and Fairness}
To test our central hypothesis that samples with high aleatoric uncertainty contribute more to algorithmic unfairness and prediction errors, we conduct additional experiments for tabular data to examine how GAIA performs in terms of utility and fairness when removing samples with high aleatoric uncertainty~(\textbf{RQ2}). Fig.~\ref{fig:pruning} shows results for pruning samples with high aleatoric uncertainty. For both Adult and German datasets, we observe improved accuracy and EOD as we filter out the most uncertain predictions. Group fairness metrics, such as EOD and AOD, consider the difference between the TPR and FPR. When the predictions completely match the ground truth, these metrics approach $1$ and $0$, respectively, for all instances of the protected group. The result is an improvement in both accuracy and fairness. Thus, if we consider the samples with the most confident predictions, the likelihood of improving both utility and fairness increases. This serves as sound empirical evidence in favor of our main hypothesis which targets the dichotomy between samples based on aleatoric uncertainty for shifting focus between fairness and utility.


\subsection{Individual Fairness}
Our fairness notion is inspired by group fairness metrics since it optimizes over the cross-entropy difference for separate instances of the protected attribute. This raises concern over its applicability for individual fairness~(\textbf{RQ3)}. However, empirical evidence from both the tabular data (Fig.~\ref{fig:dataset_method_comparison}, right) and image data (Table \ref{tab:gaia_images}) shows that GAIA also performs well on the trade-off when individual fairness metrics are of particular interest. To understand these results, we again consider the dichotomy between regions of high and low aleatoric uncertainty and the two individual fairness metrics we used (GE and CS). 

The Generalized Entropy Error (GE) is a metric that quantifies the entropy index within each group. When we have low aleatoric uncertainty within a single group, the predictor tends to closely match the ground truth for each sample. This is because higher confidence increases the likelihood of a prediction aligning with the actual label. On the other hand, when the aleatoric uncertainty is high, GAIA aims to optimize for equal cross-entropy between groups, which contributes to improved fairness. However, it is important to note that in scenarios where aleatoric uncertainty is high, the labels themselves are inherently noisy. Consequently, the predictive output for each sample tends to be closer to a random assignment. Thus, at an individual level, samples are treated equally. The Consistency Score (CS) is a metric that evaluates how a classifier treats its $k$ nearest neighbors. In essence, it quantifies the impact of high aleatoric uncertainty, which signifies increased variability among the labels of neighboring samples. As this noise is considered theoretically irreducible, our hypothesis is that leveraging aleatoric uncertainty can effectively identify areas where consistency can be enhanced. This approach offers insights into the improved empirical performance observed in relation to this metric.




\subsection{Summary} 
Since prior works focused on epistemic uncertainty, we study the connection between aleatoric uncertainty and fairness. We show how our approach compares against  both group and individual fairness. The results complement the findings by \textcite{binns2020apparent}, who suggest that group and individual fairness may not always be conflicting objectives. Our experiments also suggest that ALS introduces an improvement over LS. In addition, we observe that GAIA outperforms BNN consistently in terms of utility, while the BNN has a minuscule advantage in terms of fairness. BNN has a coherent representation due to the regularization effect of the variational inference on the encoding space, where the encoder must output a probabilistic distribution over the latent variables that approximates the true posterior. This encourages similar samples to have similar encodings, leading to a more organized and smoother latent space representation. Therefore, it is not surprising that BNNs demonstrate high performance on individual fairness metrics, as they evaluate the consistency in the treatment of similar covariates. Both GAIA and BNN outperform baseline approaches consistently in terms of the fairness-utility trade-off. Results over both image and tabular datasets show the generalizability of GAIA. Different architectures can be plugged in and sampling from a distribution over the model weights can be used to measure uncertainty.

\section{Related Work}
\label{sec:related}
Current work on fairness ML relies on identifying and mitigating spurious correlations or reducing epistemic uncertainty. 
We highlight the novelty of our approach in comparison. 
\subsection{Bias Mitigation and Fairness}
There are three main types of methods for reducing bias in machine learning, which depend on where in the model training process they are applied: (i) pre-processing, (ii) in-processing, and (iii) post-processing. In addition, there are various metrics for evaluating fairness that can be grouped into group fairness or individual fairness metrics. Preprocessing methods~\cite{kamiran2012data, zemel2013learning, feldman2015certifying, calmon2017optimized,cheng2022debiasing} aim to reduce bias by modifying the data, labels, or sample importance in the dataset. For example, the Disparate Impact Remover~\cite{feldman2015certifying} technique attempts to adjust the label distribution to ensure that protected attributes have the same median outcome. The Learning Fair Representations (LFR)~\cite{zemel2013learning} approach creates a latent representation of the data to obscure protected attributes. In-processing methods~\cite{agarwal2018reductions, zhang2018mitigating,cheng2021mitigating} rely on the model architecture to achieve fairness. Adversarial Debiasing~\cite{zhang2018mitigating} involves an adversary that tries to predict the protected attribute. The goal is to make the best predictions in a way that prevents the adversary from distinguishing the protected attribute. Post-processing methods~\cite{kamiran2012decision, hardt2016equality, pleiss2017fairness} adjust the predictions of a trained model after inference to make them unbiased. There are various approaches with different debiasing objectives. Some methods target specific fairness metrics, such as Calibrated Equal Odds Difference~\cite{pleiss2017fairness}, which aims to minimize Equalized Odds. 
\subsection{Uncertainty based Learning}
Deep Learning has achieved unprecedented success in making accurate predictions in various domains; therefore, it is increasingly important to evaluate the reliability and uncertainty of AI systems before deployment. The principles of uncertainty play an important role in AI settings such
as concrete learning algorithms \cite{mitchell1980need} and active learning \cite{nguyen2019epistemic}. There are two main types of uncertainty, i.e., aleatoric (or data) uncertainty and epistemic (or model) uncertainty \cite{hullermeier2021aleatoric}. Common techniques used in uncertainty quantification include Bayesian ~\cite{neal2012bayesian,wang2018adversarial} and Ensemble \cite{zhang2020mix,lakshminarayanan2017simple} methods. The highlights come in the form of popular variational inference approaches such as Variational Auto-Encoders (VAE) \cite{kingma2019introduction}. 
The specialty of VAE comes from the estimation of a distribution in the latent space rather than a specific latent representation. 
Similarly, Bayesian Neural Networks (BNNs) use a distribution over the weights, rather than specific weights to estimate the uncertainty for predictions. 
One common variation of BNNs is Bayes by Backprop~\cite{blundell2015weight} which leverages the standard backpropagation used in traditional NNs. 
Despite the popularity of uncertainty quantification, approaches using uncertainty to improve fairness are scarce. ROC~\cite{kamiran2012decision} is one such instance. \textcite{liupushing} use a multi-task model for predicting the under-represented class label in addition to the classification label to create a robust representation space. \textcite{singh2021fairness} propose an approach for fair ranking where the probability of being assigned a higher rank is in proportion to the estimated merit. 

Our approach complements past work by incorporating aleatoric uncertainty in particular. While prior works suggest good heuristics and processing techniques to overcome the challenge of lack of data, our approach suggests that when the model is likely to make the correct prediction, it is also likely to be fair. Conversely, when the model is unlikely to make the correct prediction due to data ambiguity, we optimize it to ensure fairness. Past approaches can easily be incorporated into our proposed framework by substituting them with the utility objective.

\section{Conclusion and Future Work}
\label{sec:conclusion}

This study introduces a novel concept balancing fairness and utility via aleatoric uncertainty. By optimizing objectives based on uncertainty levels, our approach improves fairness and utility trade-off. Aleatoric uncertainty informs model decisions for better trade-off. To mitigate the confounding effects associated with protected attributes, we propose a distributional intervention approach when estimating uncertainty using BNN. We then optimize for fairness in the solution space with high aleatoric uncertainty, and utility elsewhere. The proposed GAIA approach yields an improved fairness-utility trade-off regarding both group and individual fairness. A thorough evaluation of our approach is conducted using multiple datasets across various domains, various metrics, and comparisons to established baseline methods. The theoretical analyses and empirical evidence provide insights into the advantages, limitations, and areas for further improvement in our concept.

Our work significantly contributes to the field of ML by offering a new solution to the balance between fairness and utility. The study highlights the potential link between fairness and predictive uncertainty, and future research will delve into the robustness, scalability, and potential applications of this concept in other domains. 

While our approach demonstrates promising results, we acknowledge a few limitations. GAIA relies on the differences in uncertainty between training samples. If the majority of samples consistently exhibit low uncertainty, it suggests both high utility and fairness, even with simple approaches that do not specifically focus on fairness, such as Empirical Risk Minimization~\cite{vapnik1991principles}. However, if most samples consistently exhibit high uncertainty, our training objective leans toward maintaining fairness rather than utility.

Altering the uncertainty quantification backbone architecture, such as using an auto-encoder, could provide additional insights, and our design allows for such modifications. We separate the downstream model from the uncertainty model, enabling easy integration of GAIA with existing architectures for downstream tasks.

\section*{Acknowledgments}
This work received support from the National Science Foundation (NSF) under grant number 2036127, as well as from the Cisco Research Gift Grant (Lu Cheng). The opinions, interpretations, conclusions, and recommendations presented herein solely reflect those of the authors.



\printbibliography

@inproceedings{feldman2015certifying,
  title={Certifying and removing disparate impact},
  author={Feldman, Michael and Friedler, Sorelle A and Moeller, John and Scheidegger, Carlos and Venkatasubramanian, Suresh},
  booktitle={proceedings of the 21th ACM SIGKDD international conference on knowledge discovery and data mining},
  pages={259--268},
  year={2015}
}

@inproceedings{zemel2013learning,
  title={Learning fair representations},
  author={Zemel, Rich and Wu, Yu and Swersky, Kevin and Pitassi, Toni and Dwork, Cynthia},
  booktitle={International conference on machine learning},
  pages={325--333},
  year={2013},
  organization={PMLR}
}

@inproceedings{cheng2021mitigating,
  title={Mitigating bias in session-based cyberbullying detection: A non-compromising approach},
  author={Cheng, Lu and Mosallanezhad, Ahmadreza and Silva, Yasin N and Hall, Deborah L and Liu, Huan},
  booktitle={ACL-IJCNLP},
  volume={1},
  year={2021}
}

@inproceedings{cheng2022debiasing,
  title={Debiasing Word Embeddings with Nonlinear Geometry},
  author={Cheng, Lu and Kim, Nayoung and Liu, Huan},
  booktitle={Proceedings of the 29th International Conference on Computational Linguistics},
  pages={1286--1298},
  year={2022}
}

@article{abdar2021review,
  title={A review of uncertainty quantification in deep learning: Techniques, applications and challenges},
  author={Abdar, Moloud and Pourpanah, Farhad and Hussain, Sadiq and Rezazadegan, Dana and Liu, Li and Ghavamzadeh, Mohammad and Fieguth, Paul and Cao, Xiaochun and Khosravi, Abbas and Acharya, U Rajendra and others},
  journal={Information Fusion},
  volume={76},
  pages={243--297},
  year={2021},
  publisher={Elsevier}
}

@article{geirhos2020shortcut,
  title={Shortcut learning in deep neural networks},
  author={Geirhos, Robert and Jacobsen, J{\"o}rn-Henrik and Michaelis, Claudio and Zemel, Richard and Brendel, Wieland and Bethge, Matthias and Wichmann, Felix A},
  journal={Nature Machine Intelligence},
  volume={2},
  number={11},
  pages={665--673},
  year={2020},
  publisher={Nature Publishing Group UK London}
}

@article{kamiran2012data,
  title={Data preprocessing techniques for classification without discrimination},
  author={Kamiran, Faisal and Calders, Toon},
  journal={Knowledge and information systems},
  volume={33},
  number={1},
  pages={1--33},
  year={2012},
  publisher={Springer}
}

@article{hullermeier2021aleatoric,
  title={Aleatoric and epistemic uncertainty in machine learning: An introduction to concepts and methods},
  author={H{\"u}llermeier, Eyke and Waegeman, Willem},
  journal={Machine Learning},
  volume={110},
  pages={457--506},
  year={2021},
  publisher={Springer}
}

@article{cheng2021socially,
  title={Socially responsible ai algorithms: Issues, purposes, and challenges},
  author={Cheng, Lu and Varshney, Kush R and Liu, Huan},
  journal={Journal of Artificial Intelligence Research},
  volume={71},
  pages={1137--1181},
  year={2021}
}

@inproceedings{dutta2020there,
  title={Is there a trade-off between fairness and accuracy? a perspective using mismatched hypothesis testing},
  author={Dutta, Sanghamitra and Wei, Dennis and Yueksel, Hazar and Chen, Pin-Yu and Liu, Sijia and Varshney, Kush},
  booktitle={International Conference on Machine Learning},
  pages={2803--2813},
  year={2020},
  organization={PMLR}
}

@inproceedings{zhang2018mitigating,
  title={Mitigating unwanted biases with adversarial learning},
  author={Zhang, Brian Hu and Lemoine, Blake and Mitchell, Margaret},
  booktitle={Proceedings of the 2018 AAAI/ACM Conference on AI, Ethics, and Society},
  pages={335--340},
  year={2018}
}

@inproceedings{chuangfair,
  title={Fair Mixup: Fairness via Interpolation},
  author={Chuang, Ching-Yao and Mroueh, Youssef},
  booktitle={International Conference on Learning Representations}
}

@inproceedings{liu2015faceattributes,
  title = {Deep Learning Face Attributes in the Wild},
  author = {Liu, Ziwei and Luo, Ping and Wang, Xiaogang and Tang, Xiaoou},
  booktitle = {Proceedings of International Conference on Computer Vision (ICCV)},
  month = {December},
  year = {2015} 
}

@inproceedings{blundell2015weight,
  title={Weight uncertainty in neural network},
  author={Blundell, Charles and Cornebise, Julien and Kavukcuoglu, Koray and Wierstra, Daan},
  booktitle={International conference on machine learning},
  pages={1613--1622},
  year={2015},
  organization={PMLR}
}

@article{cao2020heteroskedastic,
  title={Heteroskedastic and imbalanced deep learning with adaptive regularization},
  author={Cao, Kaidi and Chen, Yining and Lu, Junwei and Arechiga, Nikos and Gaidon, Adrien and Ma, Tengyu},
  journal={arXiv preprint arXiv:2006.15766},
  year={2020}
}

@inproceedings{wang2018adversarial,
  title={Adversarial distillation of bayesian neural network posteriors},
  author={Wang, Kuan-Chieh and Vicol, Paul and Lucas, James and Gu, Li and Grosse, Roger and Zemel, Richard},
  booktitle={International conference on machine learning},
  pages={5190--5199},
  year={2018},
  organization={PMLR}
}

@inproceedings{zhang2020mix,
  title={Mix-n-match: Ensemble and compositional methods for uncertainty calibration in deep learning},
  author={Zhang, Jize and Kailkhura, Bhavya and Han, T Yong-Jin},
  booktitle={International conference on machine learning},
  pages={11117--11128},
  year={2020},
  organization={PMLR}
}

@article{kingma2019introduction,
  title={An introduction to variational autoencoders},
  author={Kingma, Diederik P and Welling, Max and others},
  journal={Foundations and Trends{\textregistered} in Machine Learning},
  volume={12},
  number={4},
  pages={307--392},
  year={2019},
  publisher={Now Publishers, Inc.}
}

@article{lakshminarayanan2017simple,
  title={Simple and scalable predictive uncertainty estimation using deep ensembles},
  author={Lakshminarayanan, Balaji and Pritzel, Alexander and Blundell, Charles},
  journal={Advances in neural information processing systems},
  volume={30},
  year={2017}
}

@article{mehrabi2021survey,
  title={A survey on bias and fairness in machine learning},
  author={Mehrabi, Ninareh and Morstatter, Fred and Saxena, Nripsuta and Lerman, Kristina and Galstyan, Aram},
  journal={ACM Computing Surveys (CSUR)},
  volume={54},
  number={6},
  pages={1--35},
  year={2021},
  publisher={ACM New York, NY, USA}
}

@misc{misc_adult_2,
  title        = {{Adult}},
  year         = {1996},
  author       = {UCI}, 
  howpublished = {UCI Machine Learning Repository}
}

@misc{aif360-oct-2018,
    title = "{AI Fairness} 360:  An Extensible Toolkit for Detecting, Understanding, and Mitigating Unwanted Algorithmic Bias",
    author = {Rachel K. E. Bellamy and Kuntal Dey and Michael Hind and
	Samuel C. Hoffman and Stephanie Houde and Kalapriya Kannan and
	Pranay Lohia and Jacquelyn Martino and Sameep Mehta and
	Aleksandra Mojsilovic and Seema Nagar and Karthikeyan Natesan Ramamurthy and
	John Richards and Diptikalyan Saha and Prasanna Sattigeri and
	Moninder Singh and Kush R. Varshney and Yunfeng Zhang},
    month = oct,
    year = {2018},
    url = {https://arxiv.org/abs/1810.01943}
}

@misc{german_dataset,
  author       = {Hofmann,Hans},
  title        = {{Statlog (German Credit Data)}},
  year         = {1994},
  howpublished = {UCI Machine Learning Repository}
}

@inproceedings{binns2020apparent,
  title={On the apparent conflict between individual and group fairness},
  author={Binns, Reuben},
  booktitle={Proceedings of the 2020 conference on fairness, accountability, and transparency},
  pages={514--524},
  year={2020}
}

@inproceedings{speicher2018unified,
  title={A unified approach to quantifying algorithmic unfairness: Measuring individual \&group unfairness via inequality indices},
  author={Speicher, Till and Heidari, Hoda and Grgic-Hlaca, Nina and Gummadi, Krishna P and Singla, Adish and Weller, Adrian and Zafar, Muhammad Bilal},
  booktitle={Proceedings of the 24th ACM SIGKDD international conference on knowledge discovery \& data mining},
  pages={2239--2248},
  year={2018}
}

@inproceedings{celis2019classification,
  title={Classification with fairness constraints: A meta-algorithm with provable guarantees},
  author={Celis, L Elisa and Huang, Lingxiao and Keswani, Vijay and Vishnoi, Nisheeth K},
  booktitle={Proceedings of the conference on fairness, accountability, and transparency},
  pages={319--328},
  year={2019}
}

@article{pleiss2017fairness,
  title={On fairness and calibration},
  author={Pleiss, Geoff and Raghavan, Manish and Wu, Felix and Kleinberg, Jon and Weinberger, Kilian Q},
  journal={Advances in neural information processing systems},
  volume={30},
  year={2017}
}

@inproceedings{kamiran2012decision,
  title={Decision theory for discrimination-aware classification},
  author={Kamiran, Faisal and Karim, Asim and Zhang, Xiangliang},
  booktitle={2012 IEEE 12th international conference on data mining},
  pages={924--929},
  year={2012},
  organization={IEEE}
}

@software{jax2018github,
  author = {James Bradbury and Roy Frostig and Peter Hawkins and Matthew James Johnson and Chris Leary and Dougal Maclaurin and George Necula and Adam Paszke and Jake Vander{P}las and Skye Wanderman-{M}ilne and Qiao Zhang},
  title = {{JAX}: composable transformations of {P}ython+{N}um{P}y programs},
  url = {http://github.com/google/jax},
  version = {0.3.13},
  year = {2018},
}

@misc{oryx, title={Oryx},
 url={https://github.com/jax-ml/oryx},
 author = {Sharad Vikram et al.},
 journal={GitHub},
 year={2022}}

@inproceedings{dwork2012fairness,
  title={Fairness through awareness},
  author={Dwork, Cynthia and Hardt, Moritz and Pitassi, Toniann and Reingold, Omer and Zemel, Richard},
  booktitle={Proceedings of the 3rd innovations in theoretical computer science conference},
  pages={214--226},
  year={2012}
}

@article{hardt2016equality,
  title={Equality of opportunity in supervised learning},
  author={Hardt, Moritz and Price, Eric and Srebro, Nati},
  journal={Advances in neural information processing systems},
  volume={29},
  year={2016}
}

@article{chouldechova2017fair,
  title={Fair prediction with disparate impact: A study of bias in recidivism prediction instruments},
  author={Chouldechova, Alexandra},
  journal={Big data},
  volume={5},
  number={2},
  pages={153--163},
  year={2017},
  publisher={Mary Ann Liebert, Inc. 140 Huguenot Street, 3rd Floor New Rochelle, NY 10801 USA}
}

@book{neal2012bayesian,
  title={Bayesian learning for neural networks},
  author={Neal, Radford M},
  volume={118},
  year={2012},
  publisher={Springer Science \& Business Media}
}

@article{vapnik1991principles,
  title={Principles of risk minimization for learning theory},
  author={Vapnik, Vladimir},
  journal={Advances in neural information processing systems},
  volume={4},
  year={1991}
}

@article{singh2021fairness,
  title={Fairness in ranking under uncertainty},
  author={Singh, Ashudeep and Kempe, David and Joachims, Thorsten},
  journal={Advances in Neural Information Processing Systems},
  volume={34},
  pages={11896--11908},
  year={2021}
}

@article{lin2022survey,
  title={A survey of transformers},
  author={Lin, Tianyang and Wang, Yuxin and Liu, Xiangyang and Qiu, Xipeng},
  journal={AI Open},
  year={2022},
  publisher={Elsevier}
}

@article{calmon2017optimized,
  title={Optimized pre-processing for discrimination prevention},
  author={Calmon, Flavio and Wei, Dennis and Vinzamuri, Bhanukiran and Natesan Ramamurthy, Karthikeyan and Varshney, Kush R},
  journal={Advances in neural information processing systems},
  volume={30},
  year={2017}
}

@book{mitchell1980need,
  title={The need for biases in learning generalizations},
  author={Mitchell, Tom M},
  year={1980},
  publisher={Citeseer}
}

@inproceedings{nguyen2019epistemic,
  title={Epistemic uncertainty sampling},
  author={Nguyen, Vu-Linh and Destercke, S{\'e}bastien and H{\"u}llermeier, Eyke},
  booktitle={Discovery Science: 22nd International Conference, DS 2019, Split, Croatia, October 28--30, 2019, Proceedings 22},
  pages={72--86},
  year={2019},
  organization={Springer}
}

@inproceedings{rohfairbatch,
  title={FairBatch: Batch Selection for Model Fairness},
  author={Roh, Yuji and Lee, Kangwook and Whang, Steven Euijong and Suh, Changho},
  booktitle={International Conference on Learning Representations}
}

@inproceedings{he2016deep,
  title={Deep residual learning for image recognition},
  author={He, Kaiming and Zhang, Xiangyu and Ren, Shaoqing and Sun, Jian},
  booktitle={Proceedings of the IEEE conference on computer vision and pattern recognition},
  pages={770--778},
  year={2016}
}

@article{scarlett2019introductory,
  title={An introductory guide to Fano's inequality with applications in statistical estimation},
  author={Scarlett, Jonathan and Cevher, Volkan},
  journal={arXiv preprint arXiv:1901.00555},
  year={2019}
}

@inproceedings{agarwal2018reductions,
  title={A reductions approach to fair classification},
  author={Agarwal, Alekh and Beygelzimer, Alina and Dudik, Miroslav and Langford, John and Wallach, Hanna},
  booktitle={International Conference on Machine Learning},
  pages={60--69},
  year={2018},
  organization={PMLR}
}

@article{bernheim1987coalition,
  title={Coalition-proof nash equilibria i. concepts},
  author={Bernheim, B Douglas and Peleg, Bezalel and Whinston, Michael D},
  journal={Journal of economic theory},
  volume={42},
  number={1},
  pages={1--12},
  year={1987},
  publisher={Elsevier}
}

@inproceedings{tahir2022distributional,
  title={Distributional Shift Adaptation using Domain-Specific Features},
  author={Tahir, Anique and Cheng, Lu and Guo, Ruocheng and Liu, Huan},
  booktitle={2022 IEEE International Conference on Big Data (Big Data)},
  pages={5593--5597},
  year={2022},
  organization={IEEE}
}

@article{yang2017understanding,
  title={Understanding the variational lower bound},
  author={Yang, Xitong},
  journal={variational lower bound, ELBO, hard attention},
  volume={22},
  pages={1--4},
  year={2017}
}

@article{agarap2018deep,
  title={Deep learning using rectified linear units (relu)},
  author={Agarap, Abien Fred},
  journal={arXiv preprint arXiv:1803.08375},
  year={2018}
}

@article{kingma2014adam,
  title={Adam: A method for stochastic optimization},
  author={Kingma, Diederik P and Ba, Jimmy},
  journal={arXiv preprint arXiv:1412.6980},
  year={2014}
}

@inproceedings{maas2013rectifier,
  title={Rectifier nonlinearities improve neural network acoustic models},
  author={Maas, Andrew L and Hannun, Awni Y and Ng, Andrew Y and others},
  booktitle={Proc. icml},
  volume={30},
  number={1},
  pages={3},
  year={2013},
  organization={Atlanta, Georgia, USA}
}

@inproceedings{rezaei2021robust,
  title={Robust fairness under covariate shift},
  author={Rezaei, Ashkan and Liu, Anqi and Memarrast, Omid and Ziebart, Brian D},
  booktitle={Proceedings of the AAAI Conference on Artificial Intelligence},
  volume={35},
  number={11},
  pages={9419--9427},
  year={2021}
}

@inproceedings{liupushing,
  title={Pushing the Accuracy-Group Robustness Frontier with Introspective Self-play},
  author={Liu, Jeremiah Zhe and Dvijotham, Krishnamurthy Dj and Lee, Jihyeon and Yuan, Quan and Lakshminarayanan, Balaji and Ramachandran, Deepak},
  booktitle={The Eleventh International Conference on Learning Representations}
}

@article{lahoti2020fairness,
  title={Fairness without demographics through adversarially reweighted learning},
  author={Lahoti, Preethi and Beutel, Alex and Chen, Jilin and Lee, Kang and Prost, Flavien and Thain, Nithum and Wang, Xuezhi and Chi, Ed},
  journal={Advances in neural information processing systems},
  volume={33},
  pages={728--740},
  year={2020}
}

@inproceedings{podkopaev2021distribution,
  title={Distribution-free uncertainty quantification for classification under label shift},
  author={Podkopaev, Aleksandr and Ramdas, Aaditya},
  booktitle={Uncertainty in Artificial Intelligence},
  pages={844--853},
  year={2021},
  organization={PMLR}
}

@article{donini2018empirical,
  title={Empirical risk minimization under fairness constraints},
  author={Donini, Michele and Oneto, Luca and Ben-David, Shai and Shawe-Taylor, John S and Pontil, Massimiliano},
  journal={Advances in neural information processing systems},
  volume={31},
  year={2018}
}

@STRING{oct = "Oct."}

@STRING{csur = "ACM Computing Surveys"}

@String{Computing = "Computing" }

@String{Computer = "{IEEE} Computer" }

@String{Springer = "Springer-Verlag" }

@ArtifactSoftware{R,
    title = {R: A Language and Environment for Statistical Computing},
    author = {{R Core Team}},
    organization = {R Foundation for Statistical Computing},
    address = {Vienna, Austria},
    year = {2019},
    url = {https://www.R-project.org/},
}

\end{document}